\documentclass[lettersize,journal]{IEEEtran}
\usepackage{amsmath,amsfonts}
\usepackage{algorithmic}
\usepackage{algorithm}
\usepackage{array}
\usepackage[caption=false,font=normalsize,labelfont=sf,textfont=sf]{subfig}
\usepackage{textcomp}
\usepackage{stfloats}
\usepackage{url}
\usepackage{verbatim}
\usepackage{graphicx}
\usepackage{multirow}
\usepackage{cite}
\usepackage{booktabs}
\usepackage{color}
\definecolor{mygray}{gray}{.9}
\definecolor{myblue}{RGB}{93,80,180}
\definecolor{mygreen}{RGB}{93,173,85}

\begin{document}

\title{Center-guided Classifier for Semantic Segmentation of Remote Sensing Images}

\author{Wei Zhang,
Mengting Ma,
Yizhen Jiang,
Rongrong Lian,
Zhenkai Wu,
Kangning Cui,
and Xiaowen Ma$^*$

\thanks{Wei Zhang is with the School of Software Technology, Zhejiang University, Hangzhou 310027, China, and also with the Innovation Center of Yangtze River Delta, Zhejiang University, Jiaxing Zhejiang, 314103, China.}
\thanks{Mengting Ma is with the School of Computer Science and Technology, Zhejiang University, Hangzhou 310027, China.}
\thanks{Yizhen Jiang, Rongrong Lian and Zhenkai Wu are with the School of Software Technology, Zhejiang University, Hangzhou 310027, China.}
\thanks{Kangning Cui is with the Department of Mathematics, City University of Hong Kong, Hong Kong, China.}
\thanks{Xiaowen Ma is with the School of Software Technology, Zhejiang University, Hangzhou 310027, China, and also with the Noah's Ark Lab, Huawei, Shanghai 201206, China (e-mail: xwma@zju.edu.cn).}
\thanks{* Corresponding Author}
}



\maketitle

\begin{abstract}
Compared with natural images, remote sensing images (RSIs) have the unique characteristic. i.e., larger intraclass variance, which makes semantic segmentation for remote sensing images more challenging. Moreover, existing semantic segmentation models for remote sensing images usually employ a vanilla softmax classifier, which has three drawbacks: (1) non-direct supervision for the pixel representations during training; (2) inadequate modeling ability of parametric softmax classifiers under large intraclass variance; and (3) opaque process of classification decision. In this paper, we propose a novel classifier (called CenterSeg) customized for RSI semantic segmentation, which solves the abovementioned problems with \emph{multiple prototypes},  \emph{direct supervision under Grassmann manifold}, and \emph{interpretability strategy}.
Specifically, for each class, our CenterSeg obtains local class centers by aggregating corresponding pixel features based on ground-truth masks, and generates multiple prototypes through hard attention assignment and momentum updating. In addition, we introduce the Grassmann manifold and constrain the joint embedding space of pixel features and prototypes based on two additional regularization terms. Especially, during the inference, CenterSeg can further provide interpretability to the model by restricting the prototype as a sample of the training set.
Experimental results on three remote sensing segmentation datasets validate the effectiveness of the model. Besides the superior performance, CenterSeg has the advantages of simplicity, lightweight, compatibility, and interpretability.  Code is available at \url{https://github.com/xwmaxwma/rssegmentation}.
\end{abstract}


\section{Introduction}
Semantic segmentation of remote sensing images (RSIs) is a fundamental task in the remote sensing community for parsing and understanding RSIs obtained from satellite or aerial photography\cite{geng2023dual}, which benefits various real-world applications including environmental monitoring~\cite{envir}, urban planning~\cite{urban} and agricultural management~\cite{agri}. The key objective of RSI semantic segmentation is to segment the RSI into different regions/objects and assign each of them with a specific category (e.g., water, forests, and buildings)\cite{yang2024negative, zang2024joint}.

In the early days, some RSI segmentation methods based on specific scenes (e.g., road/building segmentation~\cite{road,building}) follow the traditional segmentation framework~\cite{pspnet,ocrnet,strudel2021segmenter,xie2021segformer,nonlocal}. They mainly focus on improving special application scenarios but frequently ignore the characteristics of RSI themselves, which limits their application scopes. Recently, generic RSI semantic segmentation methods have been widely investigated, which aim to segment multiple land cover classes~\cite{unetformer,farseg,zheng2023farseg++}.

\begin{figure*}[t]
	\centering
	\includegraphics[width=0.9\textwidth]{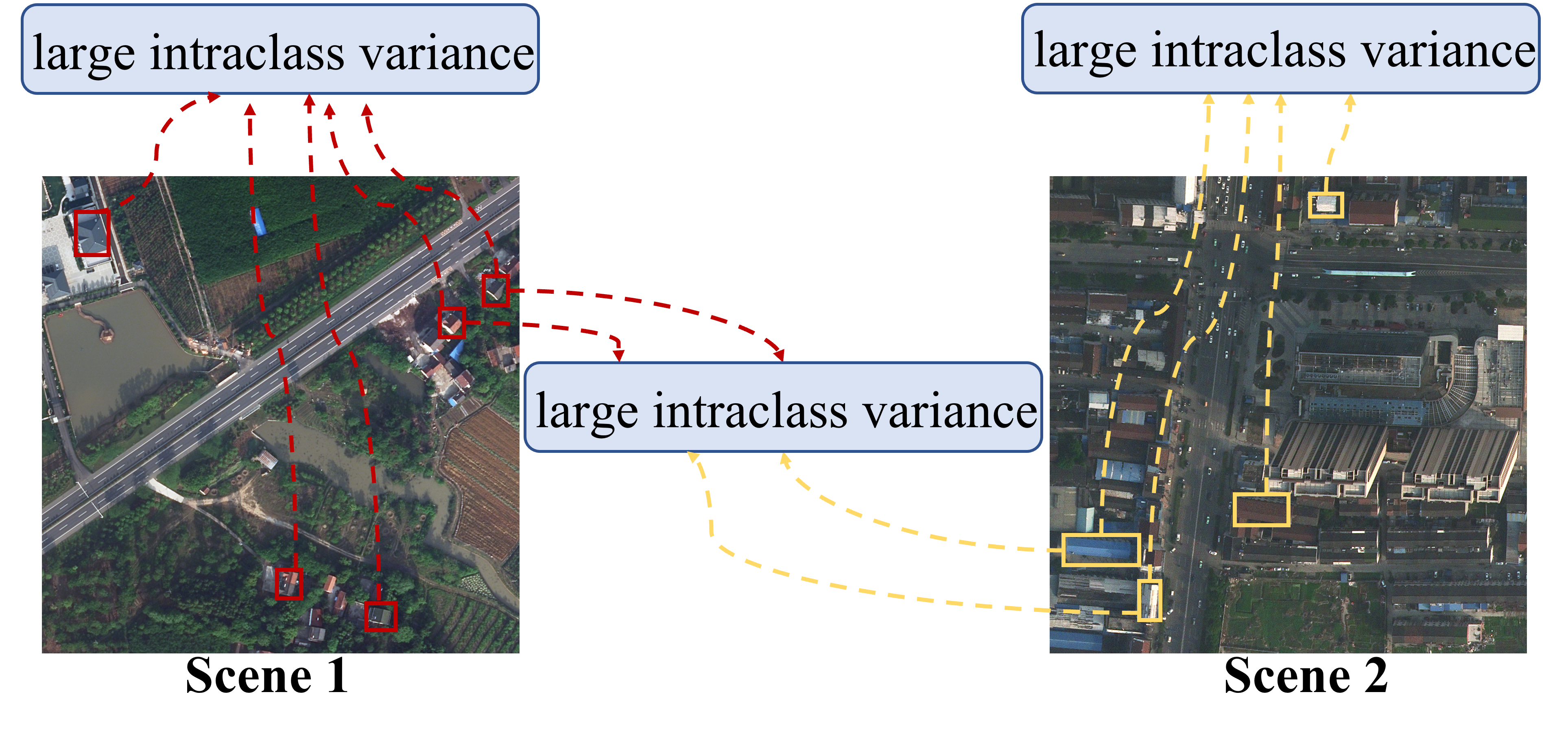}
	\caption{Display of the remote sensing image characteristic. Images are selected from the LoveDA dataset. For example, the color and shape of the building class are significantly different between Scene $1$ and Scene $2$, as well as between region and region of the same scene, underscoring the characteristic of remote sensing images, i.e., large intraclass variance.}
	\label{fig:intro}
\end{figure*}

Compared to images captured in natural scenes \cite{tinyvim}, the unique characteristic of remote sensing image makes accurate segmentation difficult, i.e., different scenes and different areas of the same scene, have significant intraclass variance due to complex spatial distributions. As illustrated in Fig.~\ref{fig:intro}, the red boxes in Scene $1$, the yellow boxes in Scene $2$, and the red and yellow boxes between the two scenes exhibit diverse colors and morphologies among themselves, yet share the same semantic (i.e., building), underscoring large intraclass variance.

A series of effective approaches have been successively proposed to mitigate the intraclass variance by optimizing backbone's design~\cite{unetformer} and context modeling~\cite{mdanet,logcan++, logcan, sacanet}. However, these RSI semantic segmentation models commonly use parametric softmax classifiers, making these approaches frequently face the following limitations: (1) parametric softmax classifiers use only one weight vector with bias to represent a specific class, which usually has large intraclass variance, which limits the performance of parametric softmax classifiers; (2) the parametric softmax classifier's cross-entropy loss function optimizes the parameters by supervising the classification probability of each pixel instead of directly supervising the pixel representation, which prevents the pixel representation from being well-trained; and (3) the weight matrix and the bias vector are both learnable parameters, where the decision-making process of their classifications is not transparent, which makes it difficult to provide interpretability for these models. 

Although some strategies~\cite{protoseg,dnc,gmmseg} have been used to improve softmax classifiers on semantic segmentation for natural images, the use of external memory and large-scale online clustering causes additional memory and computational consumption, which is suboptimal for high-resolution remote sensing images. In this paper, we propose the a center-guided classifier, called CenterSeg, customized for RSI semantic segmentation. Technically, local clustering strategy is innovatively introduced to generate representative feature samples, i.e., class centers, which can be used to accelerate model training and reduce the memory. Specifically, CenterSeg applies the ground-truth to guide feature aggregation during training, enforcing the network to generate local class centers, which are used to generate multiple semantically accurate prototypes based on hard attention assignments and momentum updates. At the inference stage, our CenterSeg directly applies these prototypes to perform classification decisions. Furthermore, To optimize the generation of prototypes, we introduce the Grassmann manifold and propose two additional regular terms to constrain the joint embedding space of pixel features and prototypes. To the best of our knowledge, our CenterSeg is the first approach to design a new classifier to improve the RSI semantic segmentation task with the following advantages:
\begin{itemize}
    \item \textbf{Effectiveness.} Due to the introduction of multiple prototypes and regularization constraints, CenterSeg is able to cope well with the large within-class variance challenge and has a good embedding space of pixel features with prototypes. We conducted extensive experiments on three RSI semantic segmentation datasets. The results show that CenterSeg can significantly improve the segmentation performance of various baseline models.

    \item \textbf{Simple.} CenterSeg does not require a lot of extra storage and computation consumption during training, which makes it easy to be applied and practically deployed.

    \item \textbf{Lightweight.} CenterSeg is tested based on the prototypes obtained during the training process, without introducing additional parameters, and the increase in computational resources is negligible.

    \item \textbf{Compatible.} CenterSeg is compatible with existing RSI semantic segmentation methods, and only needs to replace the vanilla softmax classifier.    

    \item \textbf{Interpretable.} If the prototype is restricted to samples (images) from the training set, CenterSeg can apply \emph{this looks like that} rule \cite{protopnet} to interpret its predictions, which promotes the interpretability of the model.
\end{itemize}

\begin{figure*}[t]
	\centering
	\includegraphics[width=0.99\textwidth]{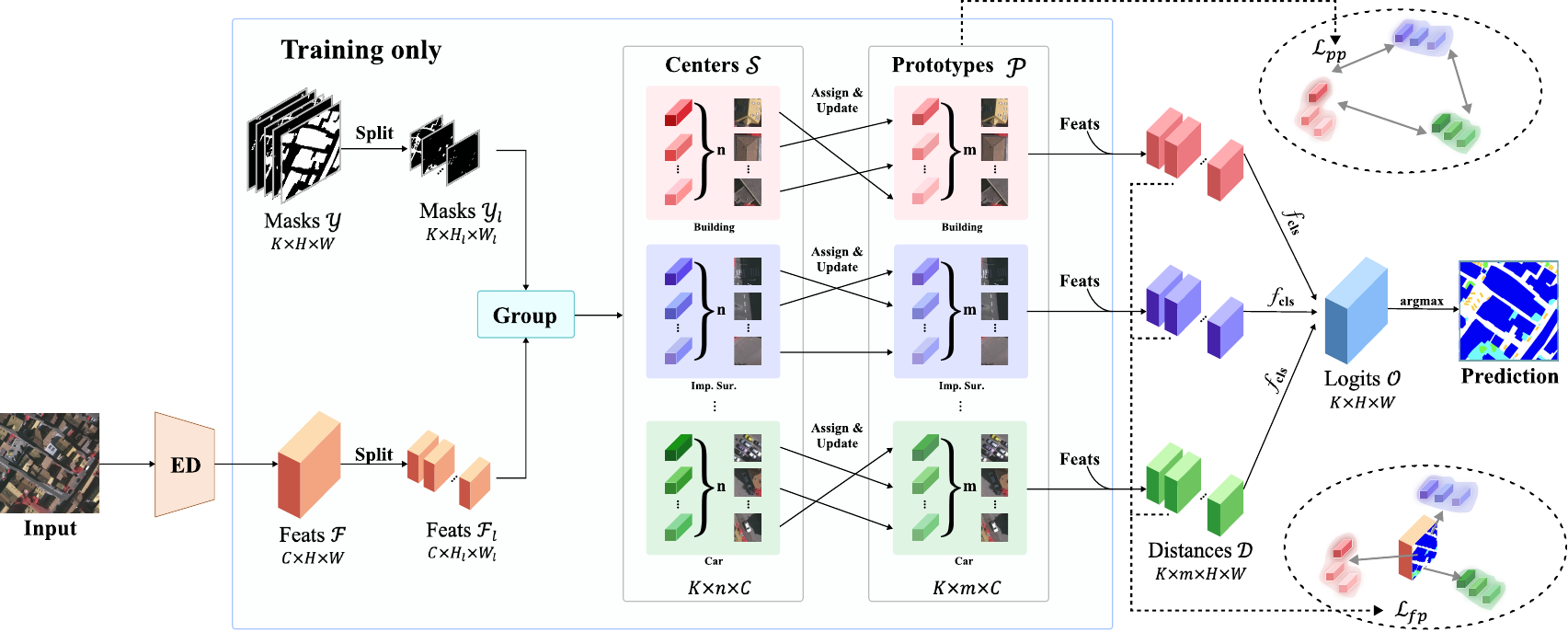}
	\caption{Architecture of the proposed CenterSeg, which consists of two key components: prototype generation and regularizer terms. At the training stage, the input image is firstly projected to a high-dimensional space after the encoder and decoder (ED) to obtain the pixel features $\mathcal{F}$. Next, we aggregate the features based on the ground-truth mask to obtain the class centers $\mathcal{S}$, i.e., the representative features of each class. Then, we generate the prototype $\mathcal{P}$ based on hard attention assignment and momentum update. In addition, two regularizer terms $\mathcal{L}_{pp}$ and $\mathcal{L}_{fp}$ are proposed to optimize the prototype generation. At the inference stage, classification decisions are performed directly based on the similarity of the pixel features to the prototype.}
	\label{fig:whole}
\end{figure*}

\section{Related Works}

\noindent \textbf{Semantic Segmentation.} Semantic segmentation is a fundamental vision task, which aims to achieve pixel-level classification of input images\cite{huang2024card}. Recently, several works focus on segmentation of RSI. Different from natural image, RSI has several unique data characteristics, i.e., complex backgrounds and large intraclass variance. To address the problem, LANet~\cite{ding2020lanet}, PointFlow~\cite{li2021pointflow}, SCO ~\cite{yang2022sparse}, and Farseg++~\cite{zheng2023farseg++} design the customized network to seek for high-quality segmentation results. However, these methods all rely on softmax classifier, which performs poorly on dataset with large intraclass variance.
In this paper, we start from the classifier perspective and design a plug-and-play classifier. As we all known, this is the first customized classifier for RSI segmentation.

\noindent \textbf{Classifier.} In response to the shortcomings of parametric softmax, several works based on new classifiers have been proposed for natural semantic segmentation tasks \cite{protoseg, ssaseg}. ProtoSeg~\cite{protoseg} first reveals the limitations of softmax classifiers in a prototypical perspective and proposes a nonparametric segmentation scheme based on unlearnable prototypes. The model represents each class as a set of unlearnable prototypes that depend only on the average features of several training pixels in the class. Dense prediction is achieved by nonparametric nearest neighbor prototype retrieval. GMMSeg~\cite{gmmseg} proposes a densely generated classifier that relies on the joint distribution of pixel features and classes prototypes. For each class, GMMSeg models the underlying data distribution by maximizing the expectation to build a Gaussian mixture model. DNC~\cite{dnc} introduces deep nearest prime, which describes the class distribution using the submasses of the training samples, and clearly interprets the classification problem as the proximity of the class submasses in the test data and the feature space. However, the above work requires external storage space for storing feature samples and the online clustering process requires significant computational resources. CenterSeg is customized for RSI segmentation tasks, which relies on the small number of classes and high intraclass variance of RSI by extracting local class centers for prototype updating. As a result, CenterSeg does not require excessive training equipment and is easier to implement and deploy.

\section{Methodology}

\textbf{Motivation.} Typical RSI semantic segmentation models \cite{zheng2023farseg++,logcan++} usually consist of three parts: an encoder, a decoder, and a classifier. Given an input image, the encoder and decoder (ED) project it to high-dimensional features $\mathcal{F} \in \mathbb{R}^{C \times H \times W}$, where $C$, $H$, and $W$ denote the channel, height, and width of it, respectively. The classifier then acts as a feature descriptor whose weights are used as decision boundaries in the high-level feature space, describing the feature distribution and performs the category judgment. Previous approaches \cite{unetformer,zheng2023farseg++,mdanet} built on a parameterized softmax classifier apply a $1 \times 1$ convolution to obtain a vector of probability distributions $\mathcal{O}_i$ for each pixel $i$, and then produce the classification result $\hat{\mathcal{O}}$ by selecting the category with the highest probability. This process is formulated as:
\begin{equation}
\label{1}
    \mathcal{O}_i = \textbf{W}^T\mathcal{F}_i + \textbf{b}, \hat{\mathcal{O}}_i = \arg\max_k(\mathcal{O}_i),
\end{equation}
where $\textbf{W} \in \mathbb{R}^{K \times C}$ and $\textbf{b} \in \mathbb{R}^{K}$is the learnable weight matrix and bias, $\mathcal{F}_i \in \mathbb{R}^{C}$ is the feature vector of pixel $i$. $\arg\max_k$ denotes the argmax operation in the category dimension. The weights of the model are then optimized based on the cross-entropy loss as:
\begin{equation}
\label{2}
    \mathcal{L} = -\frac{1}{N}\sum_{i=1}^{N}\sum_{k=1}^{K}y_i^klog(\mathcal{O}_i^k),
\end{equation}
where $N$ denotes the total number of pixels; $K$ is the number of categories; and $y_i^k$ denotes the ground-truth label of the pixel $i$. As demonstrated by Eq.1 and Eq.2, the softmax classifier represents each class only based on a single learnable weight and bias (i.e., a prototype) that cannot tolerate high intraclass variance. Furthermore, it optimizes the parameters by supervising the relative classification probability of each pixel, rather than directly supervising the pixels' semantic representations (which cannot effectively constrain the embedding space of features). This motivates us to propose our CenterSeg, a center-guided classifier, to address the above issues.

\subsection{CenterSeg overview}

This section proposes our CenterSeg, a classifier customized for the RSI semantic segmentation task. As shown in Fig.~\ref{fig:whole}, our CenterSeg aggregates pixel features based on ground-truth masks to generate class centers $\mathcal{S}$ that characterize the semantic distribution of each class. Then, based on the obtained class centers $\mathcal{S}$, we generate class prototypes $\mathcal{P}$ by using hard attention assignment and momentum updating. Finally, the similarity between the pixel features $\mathcal{F}$ and each class prototype is computed and a \emph{winner-take-all} rule is applied, i.e., the segmentation result is determined by the class to which the most similar prototype belongs. This computational procedure can be formulated as:
\begin{equation}  
\hat{\mathcal{O}} = \arg\max_{m \times K} (f_\mathrm{cls}(\|\mathcal{F}-\mathcal{P}\|_2)),
\label{eq:1}
\end{equation}  
where $\mathcal{P}$ is the generated class prototypes; $m \times K$ represents the total number of class prototypes; $\|\mathcal{F}-\mathcal{P}\|_2$ denotes computing the Euclidean distance from the feature $\mathcal{F}$ to the prototype $\mathcal{P}$, and $f_{cls}$ represents the similarity calculation function as:
\begin{equation}
    f_{cls}(t) = \frac{1}{1+\alpha t}.
\end{equation}
In addition, to optimize the generation of class prototypes, we introduce the Grassmann manifold and constrain the joint embedding space representing pixel features and class prototypes based on two additional regular terms (i.e., $\mathcal{L}_{pp}$ and $\mathcal{L}_{fp}$). Therefore, the loss function of CenterSeg is:
\begin{equation}
\mathcal{L}=\mathcal{L}_{ce}+\lambda_{pp}\mathcal{L}_{pp}+\lambda_{fp}\mathcal{L}_{fp}+\lambda_{Dice}\mathcal{L}_{Dice}.
\end{equation}
where $\lambda_{pp}$, $\lambda_{fp}$ and $\lambda_{Dice}$ are adjustable parameters. This way, multiple high-quality prototypes of each class can be obtained and further constrain the feature generation space to enhance the performance of existing RSI segmentation models \cite{farseg,unetformer}. In the following section, we describe the generation process of the prototype (Sec. \ref{sec:cpg}) and the regular term (Sec. \ref{sec:reg}) in details.

\begin{figure}[t]
	\centering
    \includegraphics[width=\columnwidth]{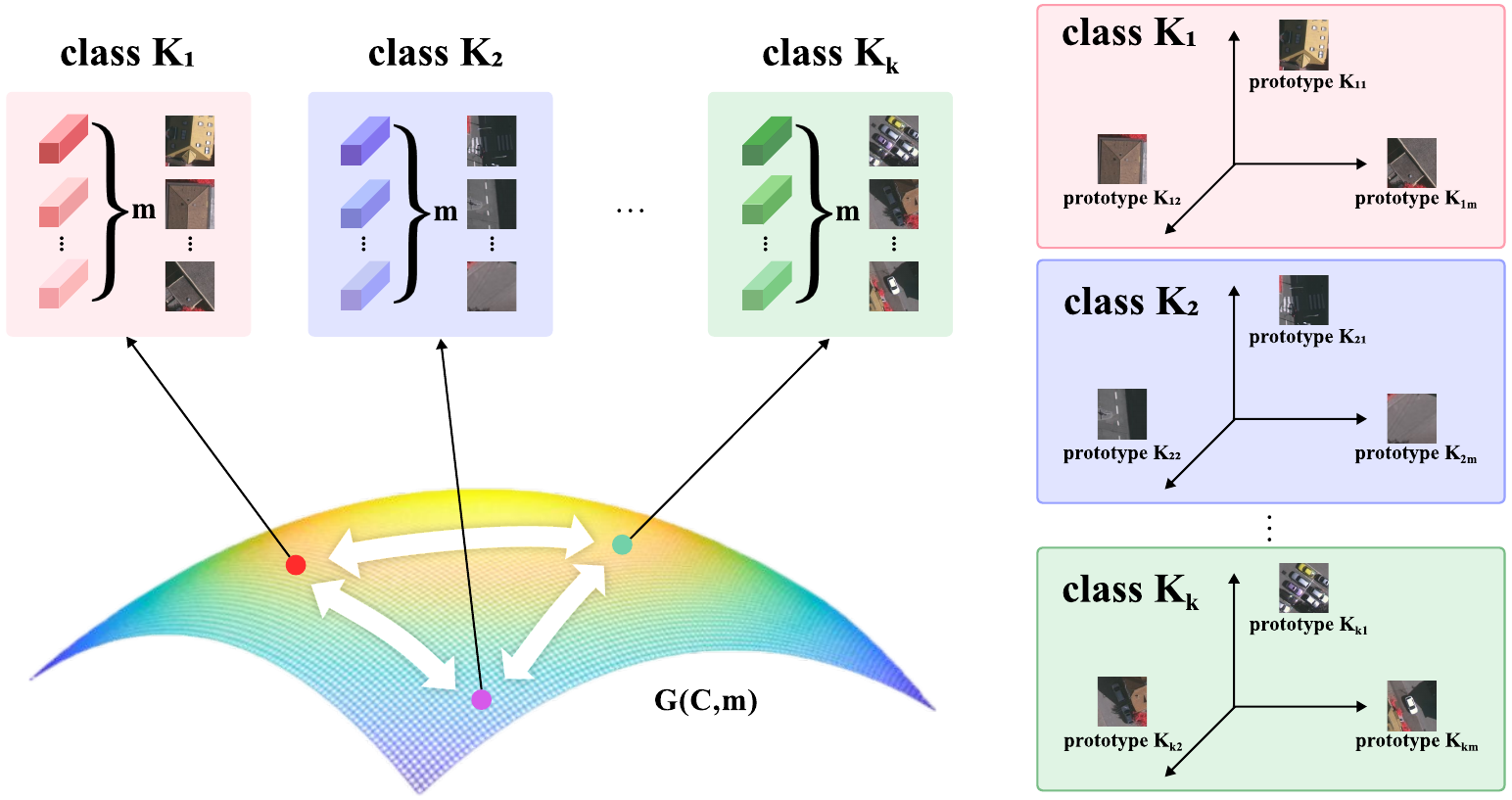}%
	\caption{Illustration of Grassmann Manifold semantic space. Left shows that the space is constructed by category-aware basis concepts, where the subspace of each class could be regarded as a point on the Grassmann manifold. Right shows that the basis of subspaces representing classes are orthogonal to each other. By minimizing the distance between the projection matrices of two points on the Grassmann manifold, subspaces of different classes can be made far away from each other, thereby achieving better class separation. }
\label{fig:grassmann}
\end{figure}

\subsection{Class Prototypes Generation}
\label{sec:cpg}

To generate semantically accurate and more discriminative prototypes, this step starts with directly aggregating pixel features based on ground-truth masks to generate class centers. Specifically, as shown in Fig. \ref{fig:whole}, it converts labels to one-hot format and then downsamples them to generate the ground-truth masks $\mathcal{Y} \in \mathbb{R}^{K \times H \times W}$ ($K$ is the number of class). To generate local class centers, the obtained $\mathcal{F}$ and $\mathcal{Y}$ are then split along the spatial dimension as: 
\begin{equation} 
\mathcal{F}_l\in \mathbb{R}^{n \times C \times (H_l\times W_l)}=\mathrm{Split}(\mathcal{F} \in \mathbb{R}^{C\times H \times W }),
\end{equation} 
\begin{equation}
\mathcal{Y}_l\in \mathbb{R}^{n \times K \times (H_l \times W_l)}=\mathrm{Split}(\mathcal{Y} \in \mathbb{R}^{K\times H \times W }),
\end{equation}
where $H_l$ and $W_l$ denote the height and width of the split patches, respectively. Consequently, the class centers $\mathcal{S}\in \mathbb{R}^{K\times n\times C}$ is obtained as:
\begin{equation}
\mathcal{S}=\mathcal{Y}_l\otimes\mathcal{R}_l,
\end{equation}
where $n = (H/H_l) \times (W/W_l)$ is the number of patches.

We then assign the obtained class centers ($\mathcal{S}$) to their corresponding class prototype via the hard attention whose similarity matrix $\mathcal{A}$ is computed via a Gumbel-softmax~\cite{gumbel} operation as:
\begin{equation}
    \mathcal{A}_k^{i,j} = \frac{\text{exp}(\mathcal{S}_k^i\mathcal{P}_k^j+\gamma_i)}{\sum_{t=1}^m\text{exp}(\mathcal{S}_k^t)\mathcal{P}_k^j+\gamma_t},
\end{equation}
where $\mathcal{S}_k^i$ and $\mathcal{P}_k^j$ denote the $i_\text{th}$ center and $j_\text{th}$ prototype of the class $k$, respectively, and $\gamma_i$ are i.i.d random samples drawn from the Gumbel (0,1) distribution. Since the hard-attention one-hot assignment operation via argmax is not differentiable, we apply the straight through trick in \cite{van2017neural} to compute the assignment matrix as:
\begin{equation}
\label{arg}
    \hat{\mathcal{A}} = \text{one-hot}(\mathcal{A}_{\text{argmax}})+\mathcal{A}-\psi(\mathcal{A}),
\end{equation}
where $\psi$ denotes the stop gradient operation. As shown in Eq.\ref{arg}, $\hat{\mathcal{A}}$ has the one-hot value of assignment to a single group, whose gradient is equal to the gradient of $\mathcal{A}$. We then perform the assignment as:
\begin{equation}
    \hat{\mathcal{P}_k^i} = \frac{1}{\mathcal{N}(i)}\sum_j\hat{\mathcal{A}}_k^{i,j} \otimes \mathcal{S}_k^j.
\end{equation}
Therefore, the prototypes of the current minibatch $\hat{\mathcal{P}}$ can be obtained, which are then used to update the final prototypes $\mathcal{P}$ in momentum,
\begin{equation}
\mathcal{P}=\mu\mathcal{P}+(1-\mu ) \hat{\mathcal{P}}.
\end{equation}
where $\mu \in[0,1]$ is the momentum update coefficient. 

\subsection{Regularizer Term}
\label{sec:reg}

To optimize the generation of prototypes, we introduce the Grassmann manifold based on two additional regular terms (i.e., $\mathcal{L}_{pp}$ and $\mathcal{L}_{fp}$), targeting at constrain the joint embedding space of pixel features and prototypes.

\subsubsection{Prototype-to-Prototype Regularizer}

\begin{figure}[t]
	\centering
        \includegraphics[width=\columnwidth]{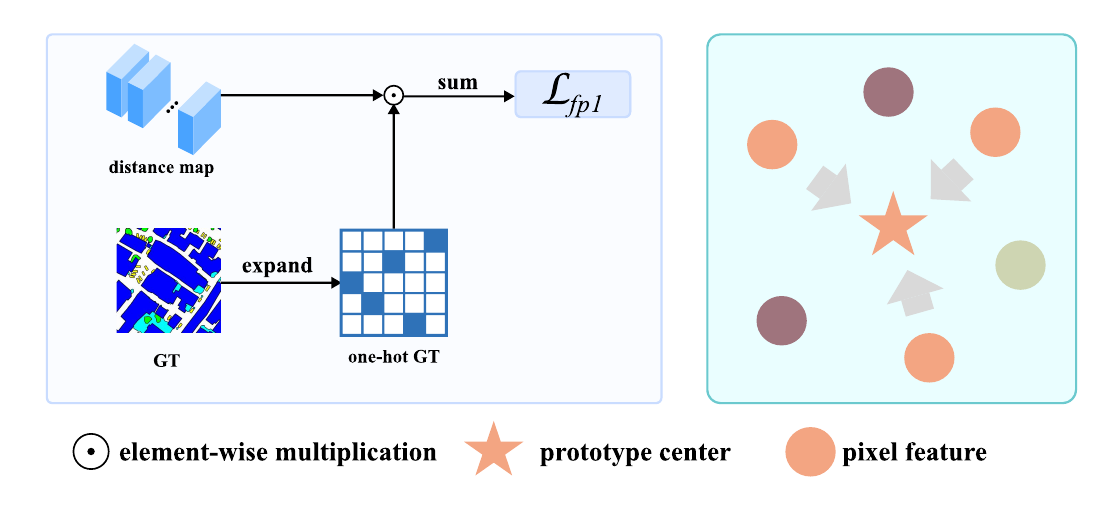}%
	\centering  
	\caption{The distance computation between pixels in $\mathcal{R}$ and the prototypes of ground-truth classes. Specifically, we apply a mask to the distance map (i.e., $\|\mathcal{F}-\mathcal{P}\|_2$) for obtaining the Euclidean distance between pixels to the corresponding prototype of class $k$, and we constrain the minimum Euclidean distance.}
\label{fig:fp1}
\end{figure}

\begin{figure}[t]
	\centering
        \includegraphics[width=\columnwidth]{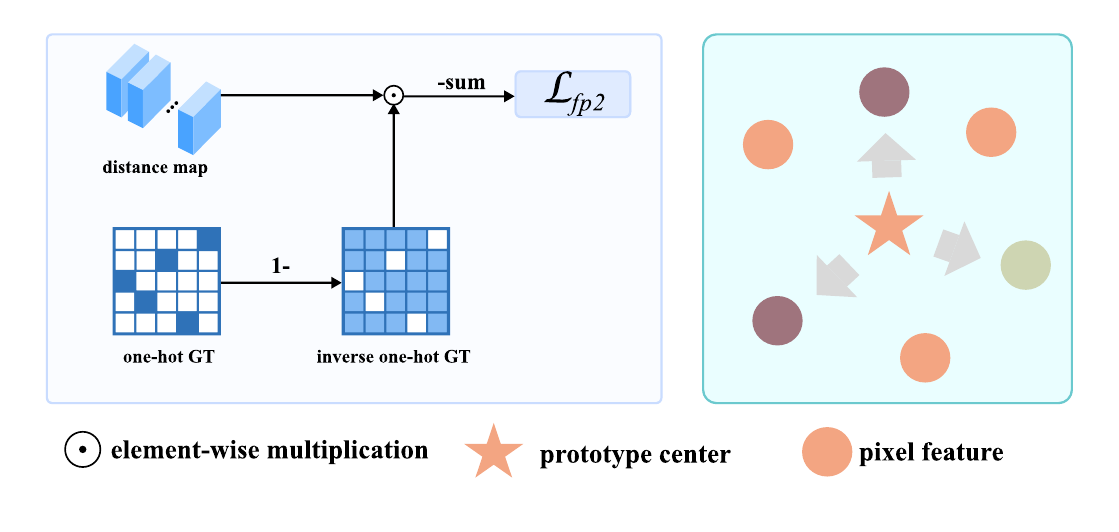}%
	\centering
	\caption{The distance computation between pixels in $\mathcal{R}$ and the prototypes of non-groundtruth classes. Specifically, we constrain the minimum distance from a pixel to a prototype of other class and hope it to be as large as possible.}
\label{fig:fp2}
\end{figure}

Our goal is to obtain diverse class prototypes which favors tolerance of large intraclass variance. This means each class is represented by a set of prototypes, each of which should have its own unique representation. Thus, all sub-prototypes of each class can be regarded as a set of basis vectors that should not be overlapped, i.e., they need to be orthogonal. To achieve this, we introduce an orthogonal idea to design the prototype-to-prototype loss $\mathcal{L}_{pp}$ based on the Grassmann Manifold semantic space (i.e., this space has good orthogonal properties).

\textbf{intraclass prototype orthogonality:} the prototype-to-prototype regularizer introduces orthogonal constraints to each prototype intraclass, thereby preventing the class prototype $\mathcal{P}$ degeneration from $m \times K$ class prototypes to $K$ class prototypes:
\begin{equation}
\mathcal{L}_{{pp}_1} = \frac{1}{K} \sum_{k=1}^{K} \left\|\mathcal{P}_k^T \mathcal{P}_k - I_m \right\|^2_F,
\end{equation}
where $\mathcal{P}_k \in \mathbb{R}^{C\times m}$ is the sub-prototypes of $k_\textit{th}$ class; $\ |\cdot\|_F$ represents $Frobenius$ norm and $I_m$ is the identity matrix.

\textbf{Orthogonalized the subspaces of different classes:} as shown in Figure~\ref{fig:grassmann}, each point on the Grassmann manifold corresponds to a unique projection matrix. By using proper projection function, vectors in different subspaces can be projected onto the corresponding subspace to obtain representation in the subspace~\cite{helmke2007newton}. Inspired by this, the subspace of each class could be regarded as a point on the Grassmann manifold in CeterSeg. In Grassmann manifolds, a common subspace distance metric is the projection metric, which can be expressed as:
\begin{equation}
\phi(\mathcal{P}_{k1},\mathcal{P}_{k2})=\frac{1}{\sqrt{2}}\left\| E(\mathcal{P}_{k1})E(\mathcal{P}_{k1})^T-E(\mathcal{P}_{k2})E(\mathcal{P}_{k2})^T \right\| _F, 
\end{equation}
By minimizing the distance between the projection matrices of two points on the Grassmann manifold, subspaces of different classes can be made far away from each other, thereby achieving better class separation. Thus, the loss function can be defined by maximizing the metric of the projection between each pair of subspaces,
\begin{equation}
\mathcal{L}_{pp_2}=\sum_{k1=1}^{K-1}\sum_{k2=k1+1}^{K}\phi(\mathcal{P}_{k1},\mathcal{P}_{k2}).
\end{equation}

As a result, the final prototype-to-prototype loss function ($\mathcal{L}_{{pp}}$) can be expressed as:
\begin{equation}
\mathcal{L}_{pp}=\mathcal{L}_{{pp}_1}+\mathcal{L}_{{pp}_2}.
\end{equation}
Therefore, the diversity of intraclass prototypes and distantness of inter-class prototypes are obtained by $\mathcal{L}_{\text{pp}}$, which facilitates to construct a solid prototype representation space.

\begin{table*}[t]
\caption{Comparative Results on the ISPRS Vaihingen and ISPRS Potsdam Datasets. Per-class best performance is marked in bold.}
\label{tab:1}
\centering
\footnotesize
\setlength{\tabcolsep}{3.2pt}{
\begin{tabular}{l | c c c c c c c c | c c c c c c c c c}
\toprule
\multirow{2}{*}{Method} & \multicolumn{8}{c|}{ISPRS Vaihingen dataset} & \multicolumn{8}{c}{ISPRS Potsdam dataset}\\ 
&  Imp. &Bui. & Low. &Tree & Car & mIoU & $\mathrm{F}_1$ &OA  &  Imp. &Bui. & Low. &Tree & Car  & mIoU & $\mathrm{F}_1$ & OA\\
\midrule \midrule
FarSeg++\scriptsize\textcolor{mygreen}{(TPAMI'23)} &86.47& 90.86& 74.39& 82.87& 76.88& 82.29 &89.77& 90.91 &87.77&93.88 &77.21&79.64&92.65&86.23 &91.99& 91.22\\
PEM\scriptsize\textcolor{mygreen}{(CVPR'24)} &85.88& 91.98& 73.65& 82.55& 76.98& 82.21&90.21 &91.22 &86.86& 93.68& 77.34& 79.65&90.23&85.55&91.73&90.48\\
LOGCAN++\scriptsize\textcolor{mygreen}{(TGRS'25)} &88.33 &91.65& 77.94& 81.68&79.85&83.89&90.87&91.85&  88.16&94.65&81.55&80.68&92.81&87.57&93.11&91.48\\
\midrule
FarSeg\scriptsize\textcolor{mygreen}{(CVPR'20)} &84.10 & 89.27 &71.10 &80.10 &71.1 &79.14&87.88  &89.57 & 85.28 & 91.33 &76.46 &77.36 &91.37 &84.36 &91.21  &89.87 \\

+ CenterSeg &\bf86.68 &\bf91.76 &\bf74.12 &\bf83.33 &\bf80.79 &\bf83.34&\bf90.80  &\bf91.20  &\bf88.14 &\bf94.39 &\bf78.63 &\bf80.76 &\bf93.43 &\bf87.07&\bf92.96  &\bf91.54 \\
\midrule
ISNet\scriptsize\textcolor{mygreen}{(ICCV'21)}  &85.91 & 90.89 &72.96 &81.93 &80.01 &82.36&90.19  &90.52 & 87.53 & 93.68 &78.60 &80.03 &93.06 &86.58&92.67  &91.27 \\

+ CenterSeg &\bf86.53 &\bf91.72 &\bf74.19 &\bf83.21 &\bf82.74 &\bf83.68 &\bf91.01 &\bf91.22 & \bf88.54 & \bf94.92 &\bf78.70 &\bf80.90 &\bf93.90 &\bf87.39&\bf93.14  &\bf91.82 \\
\midrule
ConvNeXt\scriptsize\textcolor{mygreen}{(CVPR'22)} &87.13 & 91.56 &73.81 &83.48 &78.37 &82.87 &90.50  &91.36 
& 88.31 & 93.31 &80.76 &82.45 &91.02 &87.17&93.03  &91.66\\
+ CenterSeg & \bf88.39 &\bf92.24 &\bf75.18 &\bf84.32 &\bf79.73 &\bf83.97 &\bf91.32 &\bf 91.68&  
\bf 88.89 & \bf94.77 &\bf81.86 &\bf83.66 &\bf92.53 &\bf88.34&\bf93.84  &\bf91.90\\
\midrule

			UNetFormer\scriptsize\textcolor{mygreen}{(ISPRS J'22)} &88.14 &91.47 &71.23 &81.46 &79.22 &82.30 &90.31 &90.82&  87.58 &94.23 &79.45 &80.18 &91.27 &86.54 &92.77 &91.46\\
	       + CenterSeg &\bf89.28 &\bf93.42 &\bf72.19 &\bf82.80 &\bf80.44 &\bf83.63 &\bf91.21 &\bf91.42&  \bf88.13 &\bf94.76 &\bf80.23 &\bf80.88 &\bf91.85 &\bf87.17 &\bf93.58 &\bf92.11\\
        \midrule
        EfficientViT\scriptsize\textcolor{mygreen}{(ICCV'23)}&87.26 &90.42 &69.14 &78.41 &77.37 &80.52 &87.56 &89.41
        &  84.37 &91.49 &79.66&77.43&88.20&84.23&90.11&89.61\\
        + CenterSeg & \bf89.96 &\bf93.44 &\bf72.37 &\bf80.65 &\bf80.13 &\bf83.31&\bf91.52 &\bf91.23& \bf89.23 &\bf95.67 &\bf81.46 &\bf80.89 &\bf92.33&\bf87.92 &\bf93.88 &\bf92.58\\
        \midrule
        DOCNet\scriptsize\textcolor{mygreen}{(GRSL'24)} &87.68&92.31 &72.44 &81.98 &85.64 &84.01 &91.19 &91.53&88.43 &94.97 &81.57 & 81.86&91.47 &87.64 &93.50 &92.02\\
        + CenterSeg &\bf88.45 &\bf92.51 &\bf72.85 &\bf83.87 &\bf86.33 &\bf84.80 &\bf91.63 &\bf91.97 
        &\bf88.98 &\bf95.56 &\bf82.01 &\bf82.66 &\bf92.50 &\bf88.34 &\bf94.03 &\bf92.43\\
        \midrule
        SCSM\scriptsize\textcolor{mygreen}{(ISPRS J'25)} &89.53 &92.01 &73.44 &82.23 &86.19 &84.68 &91.59 &92.22 
        &88.44 &94.48 &82.55 &79.76 & 93.72&87.79 &93.60 &92.13\\
        + CenterSeg &\bf90.12 &\bf92.67 &\bf74.43 &\bf83.11 &\bf86.50 &\bf85.37 &\bf91.89 &\bf92.56 
        &\bf89.11 &\bf95.47 &\bf83.40 &\bf80.23 &\bf93.99 &\bf88.44 &\bf94.23 &\bf92.54\\
			\bottomrule	
			\end{tabular}
        }
\label{Tab:comparsion1}
\end{table*}

\subsubsection{Feats-to-Prototype Regularizer}

\begin{table*}[t]
\caption{Comparison with state-of-the-art methods on the test set of the LoveDA dataset. Please note that the LoveDA dataset requires an online test to evaluate the model. Therefore, results for the $F_1$ and OA metrics are not available here. Per-class best performance is marked in bold.}
\label{tab:2}
\centering
\setlength{\tabcolsep}{12pt}{
\begin{tabular}{l | c c c c c c c c}
\toprule
Method&  Back. & Bui. & Road & Bar. & Was. & For. & Agr. & mIoU\\
\midrule \midrule
FarSeg++ \scriptsize\textcolor{mygreen}{(TPAMI'23)}&44.1&57.7&54.2&78.3&11.1&40.6&59.9&49.4\\
PEM \scriptsize\textcolor{mygreen}{(CVPR'24)}&43.3&56.7&53.2&79.4&14.3&41.7&58.3&49.6\\
LOGCAN++ \scriptsize\textcolor{mygreen}{(TGRS'25)}& 47.4 &58.4 &56.5 &80.1 &18.4&47.9 &64.8 &53.4\\
\midrule
FarSeg \scriptsize\textcolor{mygreen}{(CVPR'20)}&43.1 &51.5 &53.9 &76.6 &9.8 &43.3 &58.9 &48.2 \\
+ CenterSeg & \bf44.5 & \bf57.0 &53.1 &\bf78.8 &12.5 &\bf47.3 &\bf63.9 &\bf51.0 \\
\midrule
ISNet \scriptsize\textcolor{mygreen}{(ICCV'21)}&44.4 &57.4&58.0 &77.5 &\bf21.8 &43.9 &60.6 &51.9\\
+ CenterSeg & \bf46.3 & \bf58.6 &\bf60.0 &\bf78.8 &\bf23.4 &\bf44.5 &\bf62.3 &\bf53.4 \\
\midrule
ConvNeXt \scriptsize\textcolor{mygreen}{(CVPR'22)}& 46.9 & 53.5 &56.8 &76.1 &15.9 &47.5 &61.8 &51.2\\
+ CenterSeg  & \bf47.2 & 55.0 &55.1 &\bf79.4 &\bf16.8 &\bf48.6 &\bf64.3 &\bf52.3\\
\midrule
UNetFormer \scriptsize\textcolor{mygreen}{(ISPRS J'22)}&44.7 &58.8 &54.9 &79.6 &20.1 &46.0&62.5 &52.4\\
+ CenterSeg &\bf45.8 &\bf59.7 &\bf56.2 &\bf79.9 &\bf21.5 &\bf47.8 &\bf63.5&\bf53.5 \\
\midrule
EfficientViT \scriptsize\textcolor{mygreen}{(ICCV'23)}&42.9 &51.0 &52.8 &75.7 &4.3 &42.0 &61.2 &47.1\\
+ CenterSeg&\bf43.7 &\bf54.3 &\bf55.2 &\bf78.1 &\bf10.8 &\bf47.4 &\bf63.0&\bf50.4\\
\midrule
DOCNet \scriptsize\textcolor{mygreen}{(GRSL'24)}&46.6 &59.3 &56.9 &81.5 &21.3 &46.2 &65.2 &53.9\\
+ CenterSeg & 47.6 & \bf60.7&\bf57.9 &\bf81.6 &\bf22.5 &\bf47.8 &66.1 &\bf54.9 \\
\midrule
SCSM \scriptsize\textcolor{mygreen}{(ISPRS J'25)}&48.3	&60.4	&58.4	&80.7	&19.6	&47.6	&67.2 &54.6\\
+ CenterSeg & \bf49.4 & \bf61.2 &\bf59.4 &\bf80.9 &\bf21.0 &\bf47.9 &\bf67.9 &\bf55.4\\

\bottomrule	
\end{tabular}
}
\label{Tab:comparsion2}
\end{table*}

The prototype-to-prototype regularizer simply constraint class prototypes $\mathcal{P}$. In this section, we  
design the feats-to-prototype regularizer to constrain extracted features $\mathcal{F}$. 

A good joint embedding space of features and prototypes should satisfy two conditions: (1) The distance between pixel features and the class prototype of the ground-truth class to which it belongs should be as small as possible. (2) The distance between pixel features and the class prototype of the non-ground-truth class to which it belongs should be as large as possible. Therefore, we design according to the above two objectives in order to directly supervise the pixel representation (the parameterized softmax classifier supervises the relative category probabilities).

\textbf{Reduce the distance between pixel feature and the corresponding Prototype.} 
An intuitive idea is that we directly force the pixel to be equal to the most similar sub-prototype of the corresponding class. 
Specifically, as shown in Fig. \ref{fig:fp1}, we apply a mask to $\|\mathcal{F}-\mathcal{P}\|_2$ to obtain the Euclidean distance between pixels to the corresponding prototype of class $k$. We choose the minimum distance to constrain,
\begin{equation}
    \mathcal{L}_{fp1} = \sum_{k=1}^K\frac{1}{\mathcal{N}(k)}\sum_{i=1}^{\mathcal{N}(k)}\Gamma^k\text{min}(\|\mathcal{F}_i-\mathcal{P}_k\|_2),
\end{equation}
where $\mathcal{N}(k)$ denotes pixel numbers of class $k$. $\Gamma^k$ denotes the mask of class $k$, i.e., when pixel $i$ belongs to class $k$, the value is 1, else 0. Therefore, intraclass compactness is achieved by approximating a pixel and the closest sub-prototype of its corresponding class to be equal.

\textbf{Increase the distance between pixel feature and Prototype of other class.} Similarly, as shown in Fig. \ref{fig:fp2}, we constrain the minimum distance from a pixel to a prototype of other class and hope it to be as large as possible. We also add an additional parameter $\epsilon$ to impose supervision:
\begin{equation}
    \mathcal{L}_{fp2} = \sum_{k=1}^K\frac{1}{\mathcal{N}(k)}\sum_{i=1}^{\mathcal{N}(k)}\Lambda^k(\epsilon-\text{min}(\|\mathcal{F}_i-\mathcal{P}_k\|_2)),
\end{equation}
Where $\Lambda^k$ denotes the reverse mask of class $k$, i.e., when pixel $i$ belongs to class $k$, the value is $0$, else $1$. Thus, inter-class separability is achieved by distancing pixels from the nearest sub-prototypes of other classes.

The final feats-to-prototype loss function ($\mathcal{L}_{fp}$) can be expressed as:
\begin{equation}
\mathcal{L}_{fp}=\mathcal{L}_{{fp}_1}+\mathcal{L}_{{fp}_2}.
\end{equation}
$\mathcal{L}_{{fp}}$ facilitates the joint embedding space of constrained pixels and prototypes. It also achieves intraclass compactness and inter-class separability at the pixel level.

\section{Experiments}
\begin{figure*}[t]
	\centering
        \includegraphics[width=\textwidth]{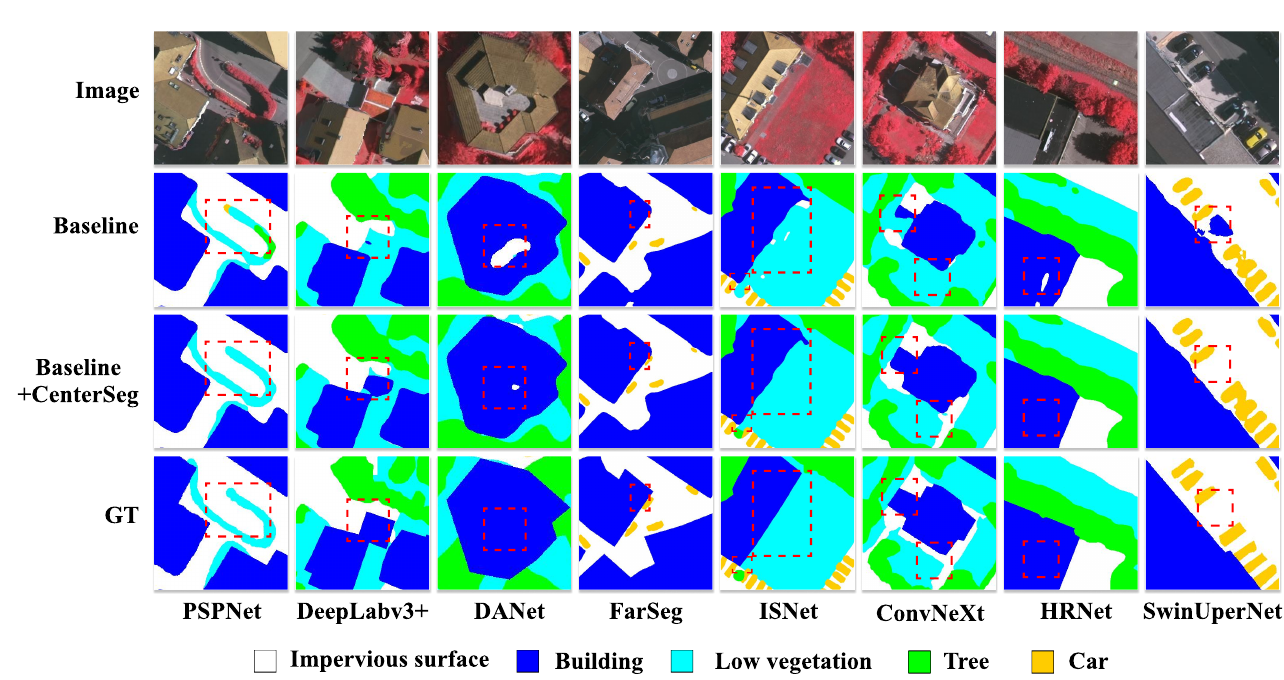}
	\centering
	\caption{Visual comparison with several baselines on the ISPRS Vaihingen test set. The red dashed box is the
area of focus. Best viewed in color and zoom in.}
\label{fig:vai_vis}
\end{figure*}

 We conduct experiments on three widely used benchmark datasets for semantic segmentation of remote sensing images, which are detailed as follows:

 \begin{figure*}[h!]
	\centering
        \includegraphics[width=\textwidth]{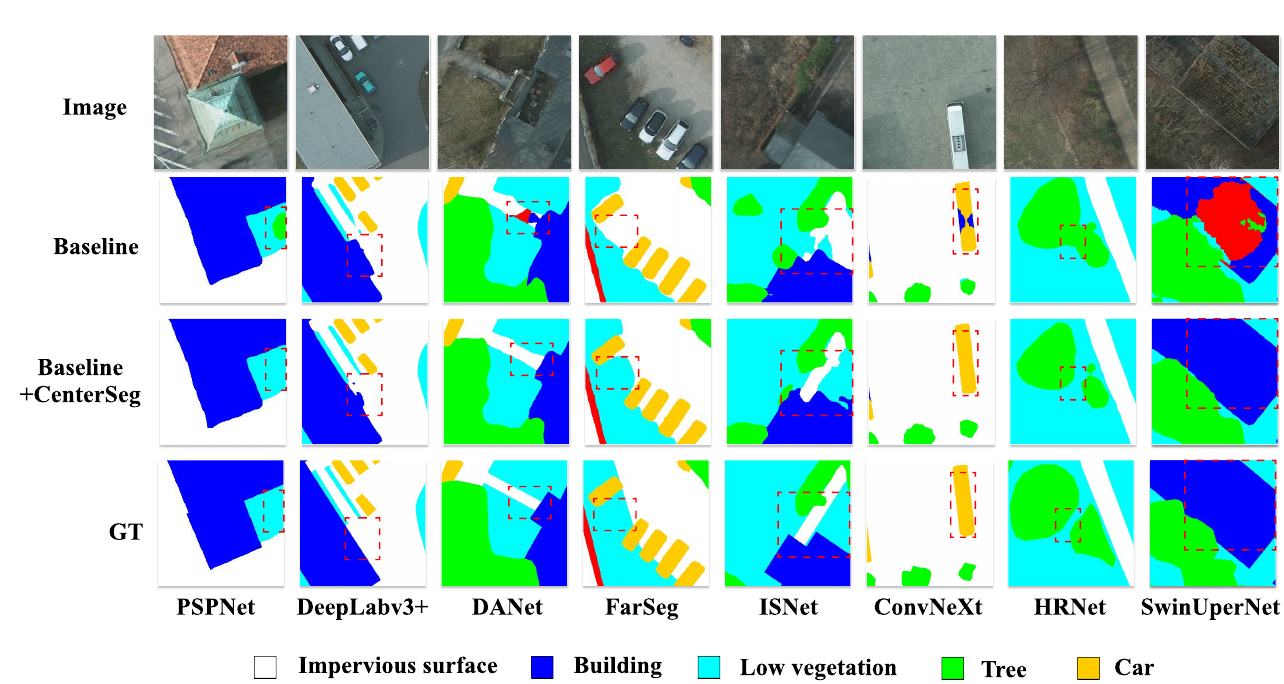}
	\centering
	\caption{Visual comparison with several baselines on the Potsdam test set. The red dashed box is the
area of focus. Best viewed in color and zoom in.}
\label{fig:potsdam_vis}
\end{figure*}

\begin{figure*}[t]
	\centering
        \includegraphics[width=\textwidth]{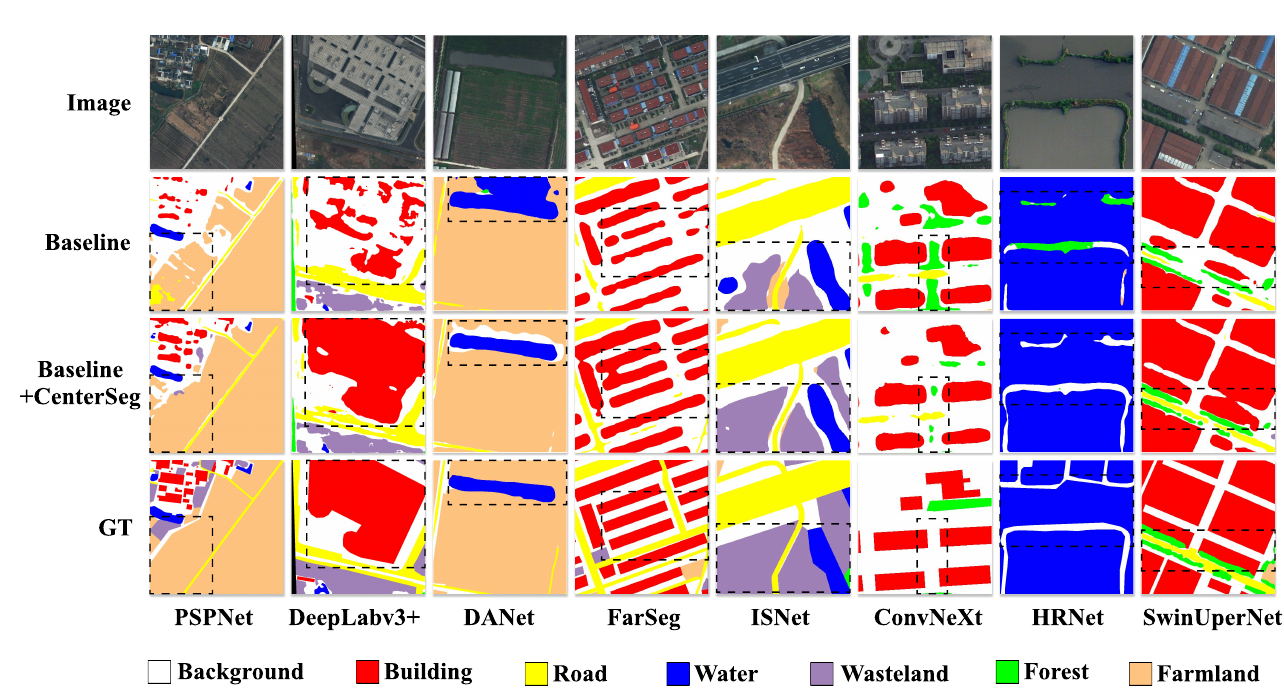}
	\centering
	\caption{Visual comparison with several baselines on the LoveDA test set. The red dashed box is the
area of focus. Best viewed in color and zoom in.}
\label{fig:loveda_vis}
\end{figure*}

\begin{figure*}[h!]
\centering \includegraphics[width=0.8\textwidth]{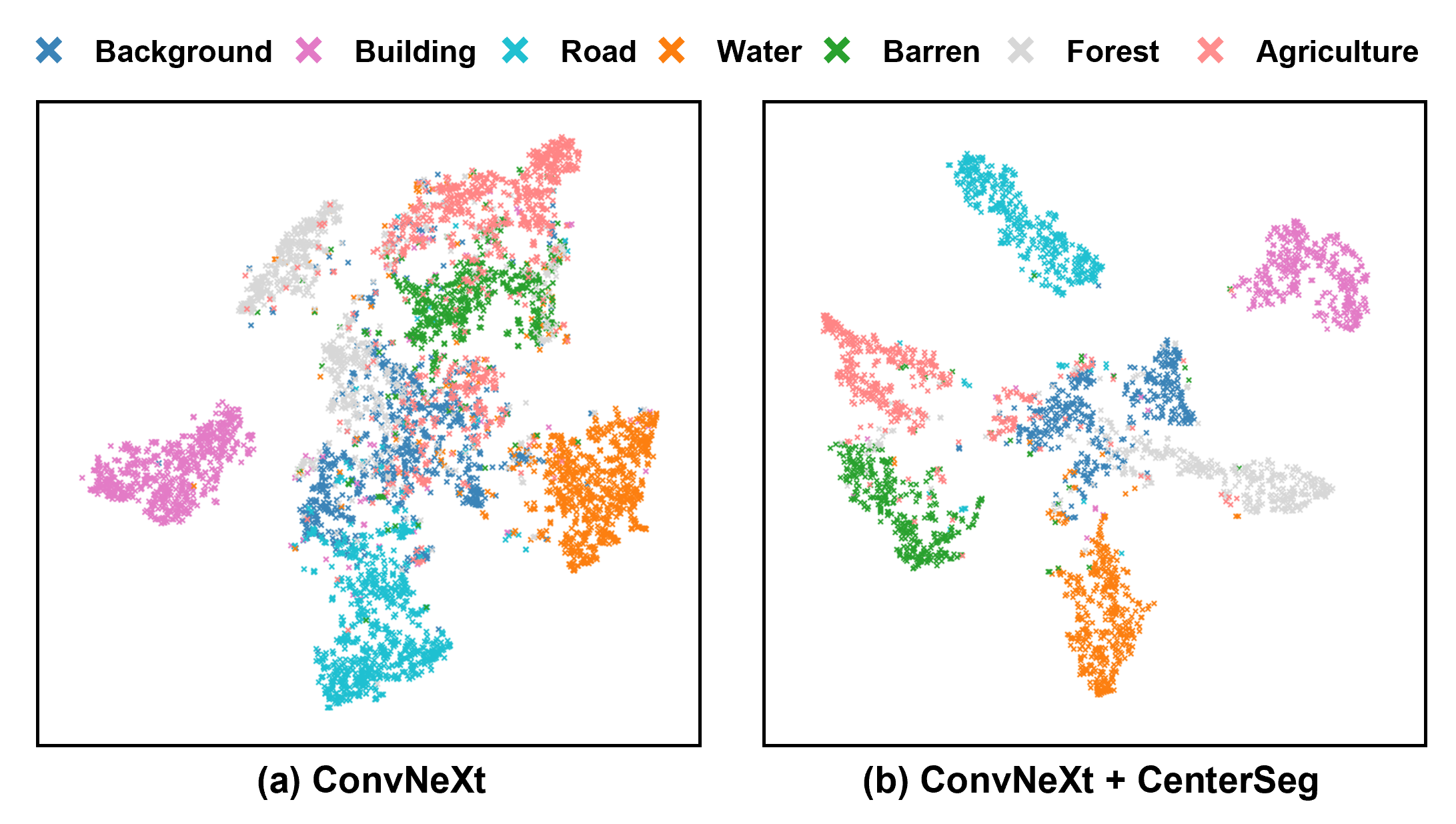}%
	\centering
	\caption{Visualization of class features distributions by the last layer of ConvNeXt and ConvNeX$+$CenterSeg. The test image is selected from LoveDA dataset. We implement the experiment with t-SNE \cite{Maaten_Hinton_2008}.}
\label{fig:3}
\end{figure*}

\subsection{Dataset}

\begin{enumerate}
\item \emph{ISPRS Vaihingen dataset} ~\cite{isprs} is a small town in southern Germany. The main building types in Vaihingen are adjacent buildings and small multistory buildings with less traffic in the town. The ground sampling distance of the Vaihingen dataset is $9$ cm. It contains $33$ image blocks, of which $16$ blocks are used for training, and the remaining $17$ blocks are used for testing. The size of each image block is different, about $2000\times2500$ pixels. Six classes of objects, namely, building, car, low vegetation, impervious surface, tree, and background, are labeled.

\item \emph{ISPRS Potsdam dataset} ~\cite{isprs} is a typical historic city in northeastern Germany. It has large buildings, narrow streets, and dense traffic. The ground sampling distance of the Potsdam dataset is $5$ cm. It contains $38$ image blocks; among them, $23$ blocks are used for training, and the remaining $14$ blocks are used for testing. The size of each image block is $6000\times 6000$ pixels. The required labeling classes are the same as the Vaihingen dataset.

\item \emph{LoveDA} dataset~\cite{loveda} intrigued a new challenge in semantic segmentation of large-scale satellite images with a spatial resolution of $0.3m$. Acquired from the Google Earth Platform, LoveDA covers a total area of over $536$ $km^2$. It includes both rural and urban areas of three cities, i.e., Nanjing, Changzhou, and Wuhan. Each image in the dataset has a spatial size of $1024\times1024$ pixels. In this study, we adopt $2522$ images for training, $1669$ images for validation, and $1796$ images for testing. The class distribution is imbalanced, and congeneric objects vary in scale, size, and surface type, making LoveDA an even more challenging dataset.
\end{enumerate}

\subsection{Evaluation Metrics} 
In this study, we use commonly adopted evaluation metrics to measure the performance of the predicted results on the testset,
including class IOU score, mean intersection over union (mIoU), overall accuracy (OA), and $\mathrm{F}_1$ score. mIoU is calculated as a global metric to evaluate accuracy. OA measures the number of correctly classified pixels against all pixels. $\mathrm{F}_1$ score is the harmonic mean of precision and recall, which evaluates the trade off between false positives and false negatives.
\begin{table*}[h!]
\caption{Comparison with existing classifiers with evaluation metrics including parameters(M), FLOPs(G), fps, Memory(GB) and mIoU.}
\label{tab:classifier}
\centering
\setlength{\tabcolsep}{10pt}{
\begin{tabular}{l | c c c c c}
\toprule
Method&  Params & FLOPs & fps & Memory & mIoU\\
\midrule \midrule
ConvNeXt\scriptsize\textcolor{mygreen}{(CVPR'22)} & 59.3 & 235 &15.7 &5.5 &51.2\\
+ ProtoSeg\scriptsize\textcolor{mygreen}{(CVPR'22)}& 59.2 &237 &15.2 &8.5 &51.8\\
+ GMMSeg\scriptsize\textcolor{mygreen}{(NeurIPS'22)}&59.2 &236 &15.4 &7.8 &51.7 \\
+ SSA-Seg\scriptsize\textcolor{mygreen}{(NeurIPS'24)}&59.3 &235 &15.6 &5.6 &51.8 \\
+ CenterSeg &59.2 &236 &15.4&5.6&52.3\\
\bottomrule	
\end{tabular}
}
\label{tab:param}
\end{table*}

\setlength{\tabcolsep}{12pt}
\begin{table}[h!]
 \caption{Effect of the patch size and prototype numbers per-class.}
\label{tab:3}
	\centering
		\begin{tabular}{cccc}
		\toprule
                patch size & m &mIoU\\
			\midrule \midrule
			 $1\times 1$ &1 &81.60\\
              $2\times 2$ &1 &78.64\\
              $2\times 2$ &4 &83.16\\
              $2\times 2$ &8 &83.97\\
              $2\times 2$ &16 &83.35\\
              $4\times 4$ &8 &82.12\\
              $8\times 8$ &8 &68.35\\
			\bottomrule
		\end{tabular}
\label{tab:number}
\end{table}

\begin{table}[h!]
\caption{Effect of the combination of loss function.}
\label{tab:4}
\centering
		\begin{tabular}{ccccccc}
		\toprule
        $\mathcal{L}_{ce}$ 
  & $\mathcal{L}_{Dice}$
  &$\mathcal{L}_{pp}$ 
    &$\mathcal{L}_{fp}$ &mIoU\\
			\midrule \midrule
		\text{\checkmark}	& \text{\checkmark} & & &82.90 \\
            \text{\checkmark}    & \text{\checkmark} &\text{\checkmark} & &83.42 \\
		 \text{\checkmark}	& \text{\checkmark}  &&\text{\checkmark}  &83.36 \\
             \text{\checkmark}  &  &\text{\checkmark} &\text{\checkmark}&83.76 \\
              \text{\checkmark}  & \text{\checkmark} &\text{\checkmark} &\text{\checkmark}   &83.97 \\
			\bottomrule
		\end{tabular}
\label{tab:los}
\end{table}

\subsection{Implementation Details}

Our model is implemented based on the PyTorch library. All experiments are conducted using one NVIDIA A6000 GPU. The training patches are randomly sampled with the patch size of $512\times 512$ and the batch size of $4$. The stochastic gradient descent (SGD) optimizer is used with the learning rate initialized to 0.01 with a weight decay of $0.0001$. 
We choose the $0.999$ as the momentum coefficient.
The model is trained for $150$ epochs on the Vaihingen dataset, $100$ epochs on the Potsdam dataset and $50$ epochs on the LoveDA dataset.


\subsection{Comparison with State-of-the-Art Methods}
The proposed CenterSeg is a plug and play classifier which could facilate the arbitrary segmentation network to optimize. The CenterSeg is employed the commonly used high-resolution RSI semantic segmentation networks, including FarSeg~\cite{farseg}, ISNet~\cite{isnet}, UNetFormer~\cite{unetformer}, DOCNet \cite{docnet} and SCSM \cite{scsm}. In addition, We also use the recent proposed Backbone, e.g., ConvNeXt~\cite{liu2022convnet}, SwinTransformer~\cite{liu2021swin,upernet}, and EfficientViT~\cite{efficientvit}. To further illustrate the effects of Centerseg, some recent state-of-the-art models, such as FarSeg++ \cite{zheng2023farseg++}, PEM \cite{pem} and LOGCAN++ \cite{logcan++}, are used only for comparison. 

\textbf{Quantitative Analysis.} The results listed in Tab.~\ref{Tab:comparsion1} and Tab.~\ref{Tab:comparsion2} show that the CenterSeg significantly improves the performance of different RSI segmentation models in three datasets. With the introduction of the CenterSeg, a high accuracy improvement can be obtained for all the models. In the Vaihingen dataset, CenterSeg is more conducive to improving the segmentation performance of small objects (e.g., tree and car). In addition, adding CenterSeg performs relatively better on the "low vegetation" in the Potsdam dataset, which lies in the fact that "low vegetation" has more characteristic representations, i,e., large intra-class variance. The LoveDA is the challenging dataset because of more obvious intraclass diversity and inter-class similarity. Table~\ref{tab:2} shows that the CenterSeg also has superior performance on the LoveDA dataset. Overall, CenterSeg achieves state-of-the-art segmentation performance on all three datasets compared to recent popular methods such as PEM \cite{pem}, FarSeg++\cite{zheng2023farseg++}, LOGCAN++ \cite{logcan++}.



\begin{figure}[t]
	\centering
        \includegraphics[width=0.48\textwidth]{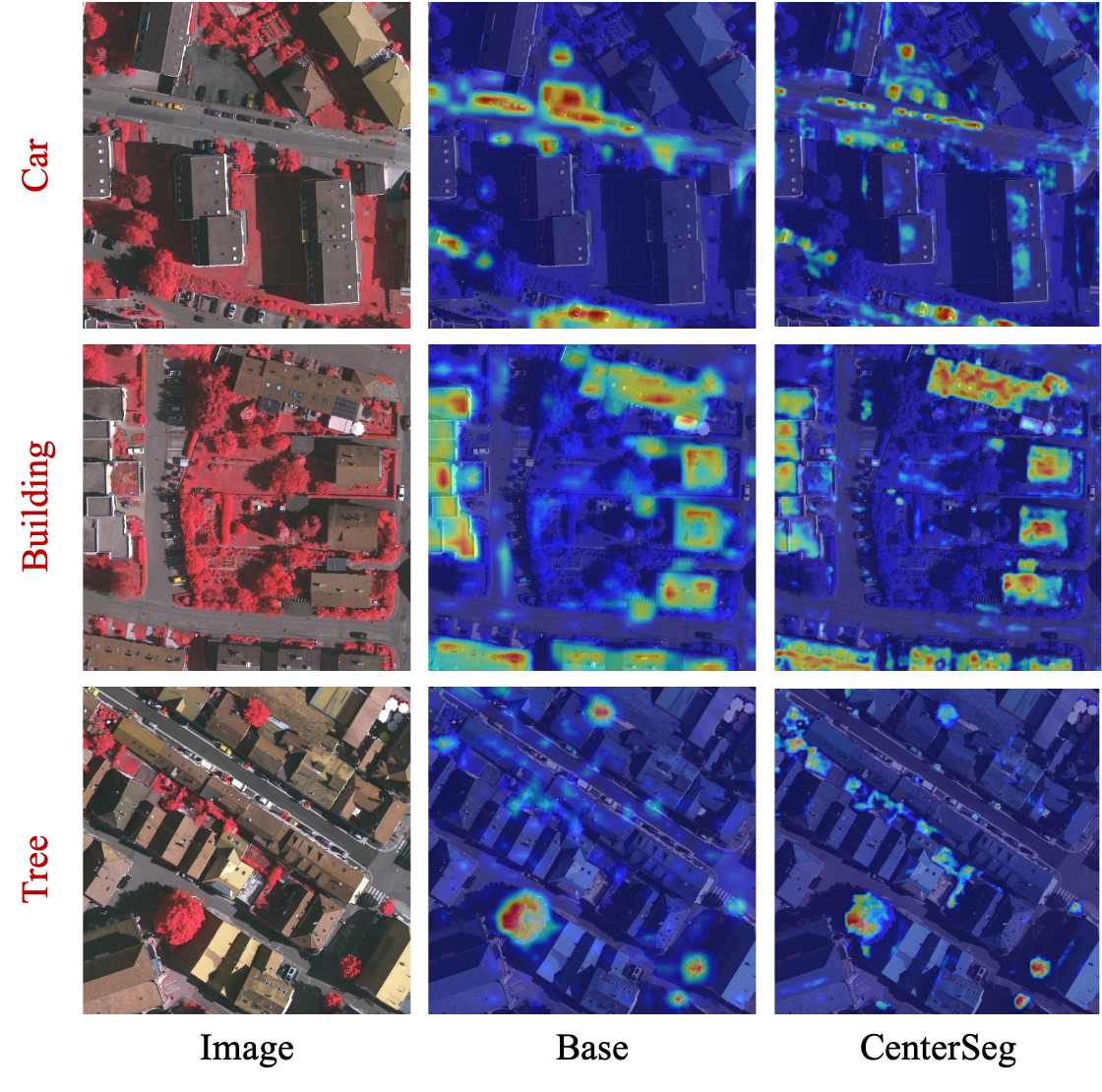}
	\centering
	\caption{Class activation maps for centerseg and baseline (ConvNeXt). Images are selected from ISPRS Vaihingen dataset. From top to bottom, the target classes are car, building and tree, respectively.}
\label{fig:pos_exp}
\end{figure}

\begin{figure*}[h!]
	\centering
        \includegraphics[width=0.9\textwidth]{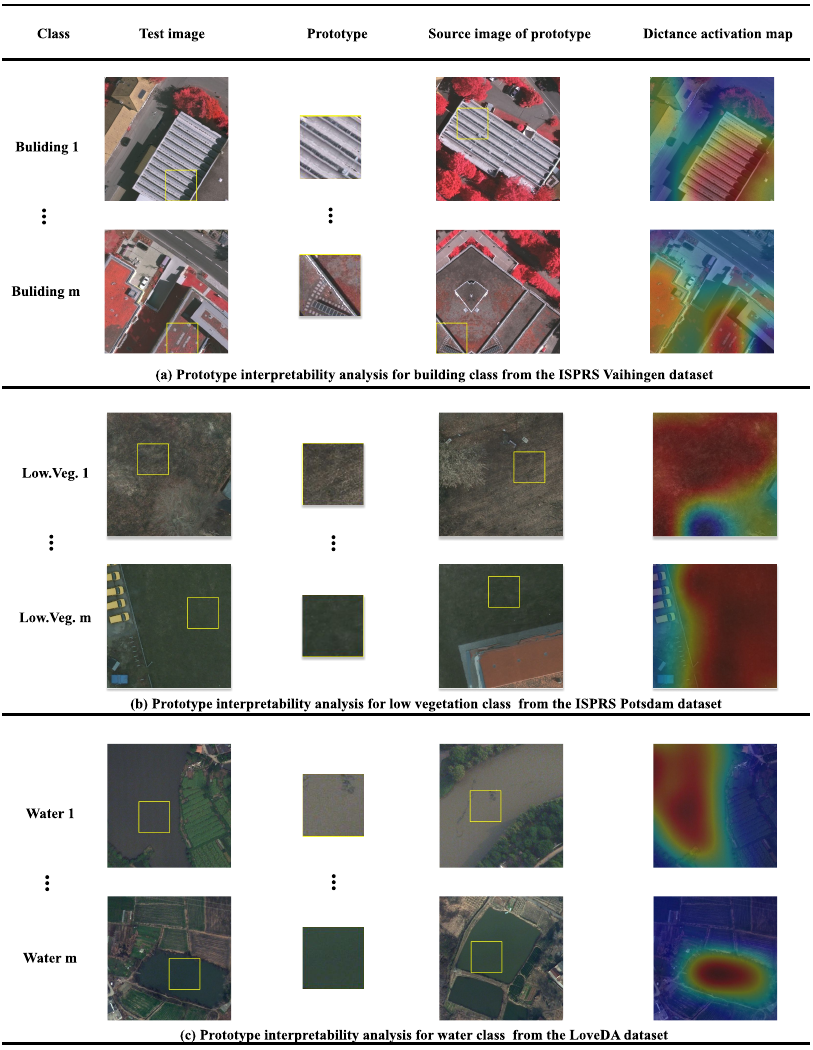}
	\centering
	\caption{Prototype interpretability analysis on three datasets. (a) Prototype interpretability analysis for buliding class from the ISPRS Vaihingen dataset. (b) Prototype interpretability analysis for low vegetation class from the ISPRS Potsdam dataset. (c) Prototype interpretability analysis for water class from the LoveDA dataset. }
\label{fig:loveda_exp}
\end{figure*}

\textbf{Visualization Analysis.} Fig. \ref{fig:vai_vis}, Fig. \ref{fig:potsdam_vis}, and Fig. \ref{fig:loveda_vis} provide additional visual comparisons on the employed three datasets, where DeepLabv3+, DANet, FarSeg, ISNet, ConvNeXt, HRNet, and UperNer are selected as the baseline models. We use dashed boxes for emphasis the focus regions. The results show that CenterSeg can obtain better visualization performance. Specifically, CenterSeg is able to fragment small objects (e.g., car classes) completely. This proves the effectiveness of CenterSeg. For some objects, such as "car" in Vaihingen dataset and "tree" in Potsdam dataset, SwinUperNet can distinguish these categories. However, several objects, such as" road" and "forest" in the LoveDA dataset, is vaguely assigned because the classes are geographically close to each other. In addition, "building" in the Potsdam dataset is incorrectly classified to "clutter" due to the extremely similar semantic features. In contrast, the SwinUperNet (CenterSeg) can obtain more local and global feature representations by adding the proposed prototype classifier. Thus, geo-objects are correctly identified whether they are natural objects  (e.g., Tree) or artificial objects (e.g., building and road). 

\textbf{Comparison with existing classifiers.}
As shown in Tab. \ref{tab:param}, we compare with existing classifiers (designed for natural images) on the LoveDA dataset. It can be observed that CenterSeg achieves better segmentation performance and efficiency, which is attributed to our customized class-centered generative multi-prototype strategy and more appropriate regularization terms for remote sensing images.

\textbf{Class Feature Distributions.}
Fig.~\ref{fig:3} illustrates the class feature distributions of ConvNeXt and ConvNeXt$+$CenterSeg by using t-SNE~\cite{Maaten_Hinton_2008}. It shows that the distinction between each class is not obvious in ConvNeXt shown in Figure~\ref{fig:3} (a). In constrast, ConvNeXt $+$ CenterSeg shows a more compact distribution in the feature space than the ConvNeXt in certain classes, e.g., forest and road.


\section{Ablation Study}

We perform ablation studies on ISPRS Vaihingen dataset. The ConvNeXt is selected as the baseline in all studies.

\textbf{Class Activation Map.} As shown in Fig. \ref{fig:pos_exp}, we set car, building and tree as the target classes to obtain the corresponding class activation map based on Grad-CAM \cite{Selvaraju_Cogswell_Das_Vedantam_Parikh_Batra_2020}. It can be observed that the CenterSeg have higher activations in the target classes, as well as concentrates more on the building/tree/car area than the Baseline (ConvNeXt).

\textbf{Prototype Interpretability.} Prototype-based classifiers are favored for their interpretability over parametric softmax classifiers. Then, there is no related research work in the field of semantic segmentation of remote sensing images. Our approach is, to some extent, inspired by ProtoPNet~\cite{protopnet}, ProtoSeg~\cite{protoseg} and GMMSeg~\cite{gmmseg}. Specifically, by approximating the prototype to an image patch on the training set, we can perform classification decisions based on \emph{this look like that} rule. This classification rule provides an intuitive interpretation mechanism for the model, helping us to understand the basis of the model's decisions. Furthermore, by observing and analyzing the prototype, we can infer the model's focus on different categories and distinguishing features, and thus explain the model's predictions.

As shown in Fig. \ref{fig:loveda_exp}, we perform interpretable validation on three datasets. As can be seen, CenterSeg captures well-characterized prototypes and can perform classification based on these prototypes. This gives CenterSeg clear decision rules and provides strong interpretability. To the best of our knowledge, there has been no work on interpretability in the field of remote sensing image segmentation.


\textbf{Number of Prototypes for Per-class.}
As shown Tab.~\ref{tab:number}, mIoU raises as the prototypes number increases, and reaches the maximum (i.e., $83.97$) when the prototype number is $8$. However, the segmentation results decrease slightly when the prototype number is $16$. The reason is significant prototype number is fixed in the certain dataset. The mIoU will no longer change significantly when the set prototype number is greater than the significant prototypes number.

\textbf{Combination of Loss Function.} As can be seen from the Tab.~\ref{tab:los}, the proposed regularization terms $\mathcal{L}_{pp}$ and $\mathcal{L}_{fp}$ have positive feedback. Specifically, $\mathcal{L}_{pp}$ has the greatest impact on the model as the orthogonality between class prototypes ensures the diversity of class prototypes (see line $2$). In addition, $\mathcal{L}_{fp}$ have a significant improvement in the baseline model because of increasing the intraclass compactness of the baseline (see line $3$). The model achieves the best mIoU ($83.97\%$) when using both proposed regularizer terms.

\section{Conclusion}

Considering the characteristic of remote sensing images (RSIs), i.e., large intraclass variance, we propose a novel and efficient classifier (called CenterSeg) for RSI semantic segmentation. Specifically, CenterSeg extracts local class centers for generating prototypes. In addition, to optimize the generation of prototypes, we introduce Grassmann manifold and constrain the joint embedding space of pixel features and prototypes based on two additional regular terms. Experiments demonstrate that CenterSeg has the advantages of effectiveness, simplicity, lightweight, compatibility, and interpretability, which reveal its potential for RSI segmentation tasks. We hope that this work will promote more classifier-related researches for RSI segmentation tasks.

\section{Acknowledgements}
This work is supported by the Public Welfare Science and Technology Plan of Ningbo City (2022S125) and the Key Research and Development Plan of Zhejiang Province (2021C01031).

\bibliographystyle{IEEEtran}
\bibliography{ref}




\end{document}